\documentclass[conference]{IEEEtran}
\usepackage{ifpdf}

\usepackage{cite}

\ifCLASSINFOpdf
  \usepackage[pdftex]{graphicx}
  \DeclareGraphicsExtensions{.pdf,.jpeg,.png}
\else
  \usepackage[dvips]{graphicx}
  \DeclareGraphicsExtensions{.eps}
\fi
\usepackage{amsmath}
\usepackage{algorithmic}

\usepackage{array}

\ifCLASSOPTIONcompsoc
  \usepackage[caption=false,font=normalsize,labelfont=sf,textfont=sf]{subfig}
\else
  \usepackage[caption=false,font=footnotesize]{subfig}
\fi
\usepackage{url}

\usepackage{caption2}

\usepackage{multirow}

\usepackage{amsfonts, amssymb}
\usepackage{amsthm}
\newtheorem{thm}{Theorem}

\usepackage{algorithm}

\begin{document}
%
\title{Scalable Discrete Supervised Hash Learning with Asymmetric Matrix Factorization}

\author{\IEEEauthorblockN{Shifeng Zhang, Jianmin Li, Jinma Guo, and Bo Zhang}
\IEEEauthorblockA{
State Key Lab of Intelligent Technology and Systems, TNList Lab,\\
Department of Computer Science and Technology, Tsinghua University, Beijing, 100084, China\\
{\tt \{zhangsf15@mails,lijianmin@mail,dcszb@mail\}.tsinghua.edu.cn, guojinma@gmail.com}
}
}


%


\maketitle

\begin{abstract}
Hashing method maps similar data to binary hashcodes with smaller hamming distance, and it has received a broad attention due to its low storage cost and fast retrieval speed. However, the existing limitations make the present algorithms difficult to deal with large-scale datasets: (1) discrete constraints are involved in the learning of the hash function; (2) pairwise or triplet similarity is adopted to generate efficient hashcodes, resulting both time and space complexity are greater than $O(n^2)$. To address these issues, we propose a novel discrete supervised hash learning framework which can be scalable to large-scale datasets. First, the discrete learning procedure is decomposed into a binary classifier learning scheme and binary codes learning scheme, which makes the learning procedure more efficient. Second, we adopt the {\em Asymmetric Low-rank Matrix Factorization} and propose the {\em Fast Clustering-based Batch Coordinate Descent} method, such that the time and space complexity is reduced to $O(n)$. The proposed framework also provides a flexible paradigm to incorporate with arbitrary hash function, including deep neural networks and kernel methods. Experiments on large-scale datasets demonstrate that the proposed method is superior or comparable with state-of-the-art hashing algorithms.
\end{abstract}


%
\IEEEpeerreviewmaketitle

\section{Introduction}

During the past few years, hashing has become a popular tool in solving large-scale vision and machine learning problems~\cite{li2011hashing,liu2012supervised,li2013sign}. Hashing techniques encode various types of high-dimensional data, including documents, images and videos, into compact hashcodes by certain hash functions, so that similar data are mapped to hashcodes with smaller Hamming distance. With the compact binary codes, we are able to compress data into very small storage space, and conduct efficient nearest neighbor search on large-scale datasets.

The hashing techniques are composed of {\em data-independent} methods and {\em data-dependent} methods. Locality-Sensitive Hashing (LSH) ~\cite{datar2004locality,gionis1999similarity} and MinHash~\cite{broder1998min} are the most popular {\em data-independent} methods. These methods have theoretical guarantees that similar data have higher probability to be mapped into the same hashcode, but they need long codes to achieve high precision. In contrast to {\em data-independent} hashing methods, {\em data-dependent} learning-to-hash methods aim at learning hash functions with training data. A number of methods are proposed in the literature, and we summarize them into two categories: {\em unsupervised methods}, including Spectral Hashing(SH)~\cite{weiss2009spectral}, Iterative Quantization(ITQ)~\cite{gong2013iterative}, Anchor Graph Hashing(AGH)~\cite{liu2011hashing}, Isotropic Hashing(IsoH)~\cite{kong2012isotropic}, Discrete Graph Hashing(DGH)~\cite{liu2014discrete}; and {\em supervised methods}, such as Binary Reconstructive Embeddings(BRE) ~\cite{kulis2009learning}, Minimal Loss Hashing~\cite{norouzi2011minimal}, Supervised Hashing with Kernels(KSH)~\cite{liu2012supervised}, FastHash(FastH)~\cite{lin2014fast}, Supervised Discrete Hashing(SDH)~\cite{Shen_2015_CVPR}. Experiments convey that hash functions learned by supervised hashing methods are superior to unsupervised ones.

Recent works~\cite{Shen_2015_CVPR,lin2014fast} demonstrate that more training data can improve the performances of the learned hash functions. However, existing hashing algorithms rarely discuss training on large-scale datasets. Most algorithms use pairwise or triplet similarity to learn hash functions, so that there are intuitive guarantees that similar data can learn similar hashcodes. But there are $O(n^2)$ data pairs or $O(n^3)$ data triplets where $n$ is the number of training data, which makes both the training time and space complexity at least $O(n^2)$. These methods cannot train on millions of data, like ImageNet dataset~\cite{deng2009imagenet}. Recent works like SDH~\cite{Shen_2015_CVPR} reduces the training time to $O(n)$, but it lies in the assumption that the learned binary codes are good for linear classification, thus there are no guarantees that similar hashcodes correspond to data with similar semantic information.

Moreover, the discrete constraints imposed on the binary codes lead to mix-integer optimization problems, which are generally NP-hard. Many algorithms choose to remove the discrete constraints and solve a relaxed problem, but they are less effective due to the high quantization error. Recent studies focus on learning the binary codes without relaxations. DGH~\cite{liu2014discrete} and SDH~\cite{Shen_2015_CVPR} design an optimization function in which binary constraints are explicitly imposed and handled, and the learning procedure consists of some tractable subproblems. But DGH is an unsupervised method, and SDH does not consider semantic similarity information between the training data. We consider discrete methods that can leverage the similarity information between training samples should be better for hashing.

In this paper, we propose a novel discrete learning framework to learn better hash functions. A joint optimization method is proposed, in which the binary constraints are preserved during the optimization, and the hash function is obtained by training several binary classifiers. To leverage pairwise similarity information between the training data, the similarity matrix is used in the optimization function. By making use of {\em Asymmetric Low-rank Similarity Matrix Factorization}, we reduce the computing time and storage of similarity matrix from $O(n^2)$ to $O(n)$, so our method can deal with millions of training data. To solve the most challenging binary code learning problem, we propose a novel {\em Fast Clustering-based Batch Coordinate Descent (Fast C-BCD)} algorithm to convert the binary code learning problem to a clustering problem, and generate binary codes bit by bit. We name the proposed framework as {\em \underline{Di}screte \underline{S}upervised \underline{H}ashing (DISH) Framework}.

Recent works~\cite{xia2014supervised,lai2015simultaneous,guo2016hash,zhu2016deep} show that hashing methods with deep learning can learn better hash functions. This framework is also able to learn hash functions with deep neural networks to capture better semantic information of the training data.

Our main contributions are summarized as follows:

\begin{enumerate}
    \item We propose a novel discrete supervised hash learning framework which is decomposed into a binary classifier learning scheme and binary codes learning scheme. Discrete method makes the learning procedure more efficiently.
    \item We propose the {\em Fast Clustering-based Batch Coordinate Descent} algorithm to train binary codes directly, and introduce the {\em Asymmetric Low-Rank Similarity Matrix Factorization} scheme to decompose the similarity matrix into two low-rank matrices, so that the time and space complexity is reduced to $O(n)$.
    \item The proposed DISH framework succeeds in learning on millions of data and experimental results show its superiority over the state-of-the-art hashing methods on either the retrieval performance or the training time.
\end{enumerate}

The rest of the paper is organized as follows. Section \ref{sec:related} presents the related work of recent learn-to-hash methods. Section \ref{sec:framework} introduces the {\em Discrete Supervised Hashing (DISH) Framework}, and we discuss how to combine the framework with deep neural network in Section \ref{sec:deephashing} and kernel-based methods in Section \ref{sec:kernelhash}. Experiments are shown in Section \ref{sec:exp}, and the conclusions are summarized in Section \ref{sec:conclusion}.

\section{Related Work}
\label{sec:related}
\subsection{Discrete Hashing Methods}

The goal of hash learning is to learn certain hash functions with given training data, and the hashcodes are generated by the learned hash function. Recently, many researches focus on discrete learning methods to learn hashcodes directly. Two Step Hashing (TSH)~\cite{lin2013general} proposes a general two-step approach to learn hashcodes, in which the binary codes are learned by similarities within data, and hash function can be learned by a classifier. FastH~\cite{lin2014fast} is an extension of the TSH algorithm, which improves TSH by using Boosting trees as the classifier. However, these methods learn hashcodes and hash functions separately, thus the learned binary codes may lack the relationship with the distribution of data. What's more, pair-wise similarity matrix is involved in these methods, making them not scalable. We succeed in learning hashcodes with pairwise similarity as well as achieving $O(n)$ complexity.

Some other works tried to jointly learn discrete hashcodes as well as hash functions. Discrete Graph Hashing(DGH)~\cite{liu2014discrete} designs an optimization function in which binary constraints are explicitly imposed and handled, and the learning procedure consists of two tractable subproblems. Anchor graphs are also used in this algorithm, reducing the storage of pairwise similarity matrix to $O(n)$. But it is an unsupervised algorithm and does not make use of semantic information.

Supervised Discrete Hashing(SDH)~\cite{Shen_2015_CVPR} proposes a method in which binary codes and hash functions are learned jointly and iteratively. But it cannot tackle the case where the data have no semantic labels, and it lacks theoretical and intuitive guarantees for explaining advantages of this algorithm. Moreover, it only discusses the learning of kernel-based hash functions. Our proposed framework can deal with arbitrary hash functions.

\subsection{Deep Hashing with Convolutional Networks}

Recently, deep convolutional neural network (CNN) have received great success in image classification~\cite{krizhevsky2012imagenet,simonyan2014very,he2015deep}, object detection~\cite{ren2015faster} and so on. Recent works~\cite{wan2014deep,schroff2015facenet} convey that features extracted from the last hidden layers represent more semantic information, and outperform most hand-crafted features in many vision applications.~\cite{xia2014supervised,lai2015simultaneous} show that simultaneously learning hash functions as well as the network can generate the codes with much better semantic information.

CNNH~\cite{xia2014supervised} decomposes the hash learning process into two stages. First, the approximate hashcodes are learned with pairwise similarity matrix, then the learned hashcodes are used as the supervised information to learn the deep neural network.~\cite{zhuang2016fast} use triplet loss function to generate approximate hashcodes. But the learned approximate hashcodes in these methods have no relevance with the data, making the learned nets not effective.

Some other works use one-stage method to learn binary codes and image features simultaneously.~\cite{lai2015simultaneous} uses triplet loss to learn hash function, and DHN~\cite{zhu2016deep} proposes a pairwise loss function to train the network. CNNBH~\cite{guo2016hash} is similar as SDH~\cite{Shen_2015_CVPR}, and they both assume that the learned binary codes are good for classification. For ease of back-propagation, these methods remove the discrete constraints and add some quantization penalty to decrease the quantization error. Although the penalty is introduced, the quantization error still affects the efficiency of the learned hash functions.

By incorporating our DISH framework to deep neural network, we can tackle two problems mentioned above: (1) we use the discrete method to reduce the quantization error; (2) we bridge the input data and semantic information by jointly learning hashcodes and deep neural networks.

\section{The Discrete Supervised Hashing Framework}
\label{sec:framework}

\begin{figure*}[t]
    \centering
    \includegraphics[scale=0.45]{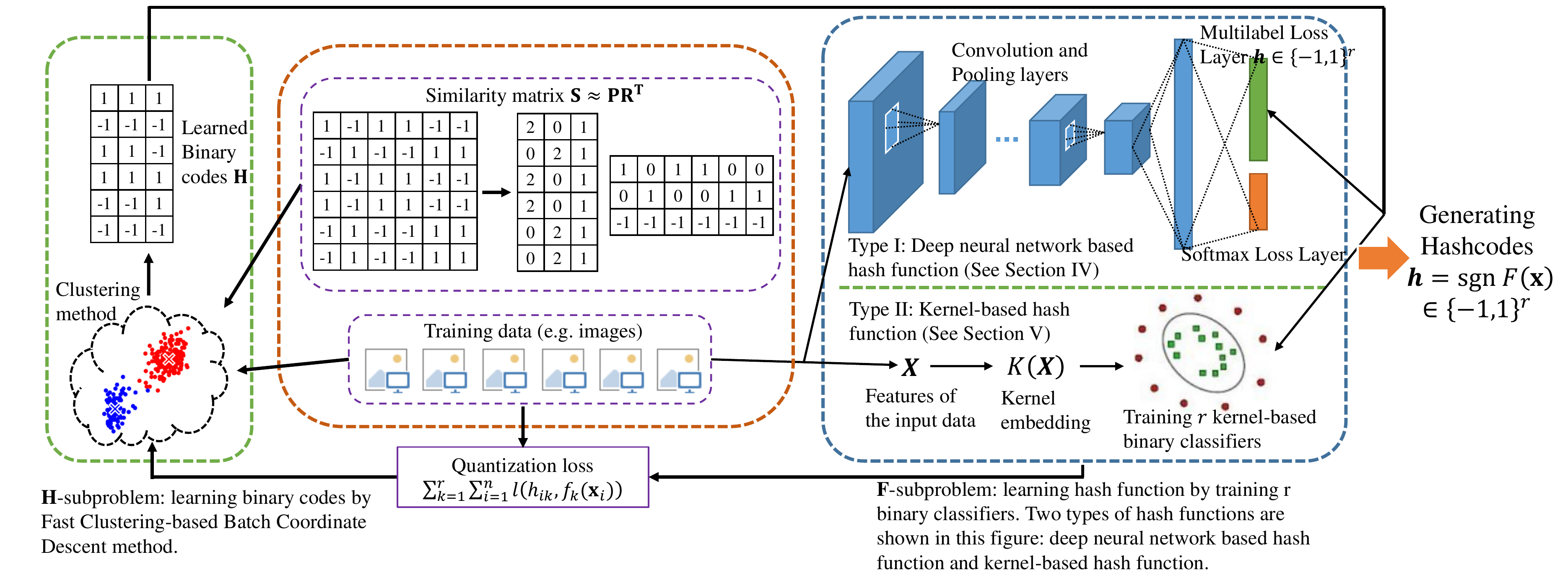}
    \caption{Overview of the {\em Discrete Supervised Hash Learning (DISH) framework}. Given input training data and similarity matrix $\mathbf{S}$, first we construct two low-rank matrices $\mathbf{P},\mathbf{R}$ to approximate $\mathbf{S}$. Then, we decompose the learning procedure into two subproblems: $\mathbf{F}$-subproblem to learn hash function by training binary classifiers, and $\mathbf{H}$-subproblem to learn discrete binary codes. Both similarity information and the training data are used in the $\mathbf{H}$-subproblem.}
    \label{fig:framework}
\end{figure*}

Suppose we are given $n$ data vectors $\mathbf{X}=[\mathbf{x}_1,\mathbf{x}_2,...,\mathbf{x}_n]^\mathrm{T}$. The goal is to learn hash function $\mathbf{H}(\mathbf{X})=[\mathbf{h}(\mathbf{x}_1),\mathbf{h}(\mathbf{x}_2),...,\mathbf{h}(\mathbf{x}_n)]^\mathrm{T} \in \{-1,1\}^{n \times r}$, where $\mathbf{h}(\mathbf{x}_i) = [h_{i1},h_{i2},...,h_{ir}]^\mathrm{T} \in \{-1,1\}^r$ is the hash function of data vector $\mathbf{x}_i$, and $r$ is the hashcode length. Denote $h_{ik}=\mathrm{sgn}(f_k(\mathbf{x}_i))$, where $\mathrm{sgn}(x)$ is $+1$ if $x>=0$ and $-1$ otherwise. Define $F(\mathbf{x})=[f_1(\mathbf{x}), f_2(\mathbf{x}),...,f_r(\mathbf{x})]^\mathrm{T} \in \mathbb{R}^{r}$ and $F(\mathbf{X})=[F(\mathbf{x}_1),F(\mathbf{x}_2),...,F(\mathbf{x}_n)]^\mathrm{T} \in \mathbb{R}^{n \times r}$, we have $\mathbf{H}(\mathbf{X})=\mathrm{sgn} (F(\mathbf{X}))$, where $\mathrm{sgn}(\cdot)$ is an element-wise sign function.

Let $\mathbf{S} = \{s_{ij}\}_{n \times n}$ be pairwise similarity matrix, in which $s_{ij}=1$ if $\mathbf{x}_i$ and $\mathbf{x}_j$ are most similar and $s_{ij}=-1$ otherwise. Then the objective to learn hash function $\mathbf{H}(\mathbf{X})$ can be formulated as
\begin{equation}
\begin{split}
\min_{F} \mathcal{Q} &= \sum_{i,j=1}^n [r s_{ij}-\mathbf{h}(\mathbf{x}_i)^{\mathrm{T}} \mathbf{h}(\mathbf{x}_j)]^2 \\
&= \Arrowvert r\mathbf{S} - \mathrm{sgn}(F(\mathbf{X}))\mathrm{sgn}(F(\mathbf{X}))^\mathrm{T} \Arrowvert_F^2
\label{obj}
\end{split}
\end{equation}
where $\| . \|_F$ is Frobenius norm of a matrix. It should be noticed that $\mathbf{h}(\mathbf{x}_i)^{\mathrm{T}} \mathbf{h}(\mathbf{x}_j)=r$ if $\mathbf{h}(\mathbf{x}_i)$ and $\mathbf{h}(\mathbf{x}_j)$ are identical and $\mathbf{h}(\mathbf{x}_i)^{\mathrm{T}} \mathbf{h}(\mathbf{x}_j) = -r$ if Hamming distance of $\mathbf{h}(\mathbf{x}_i)$ and $\mathbf{h}(\mathbf{x}_j)$ is the largest. Optimizing Eq. (\ref{obj}) means that the Hamming distance between hashcodes of similar data pairs should be small, and large otherwise.

Eq. (\ref{obj}) is hard to optimize because the sign function is involved. Inspired by~\cite{liu2014discrete}, we remove the sign function and add a quantization loss to hold the binary constraints as much as possible
\begin{equation}
\begin{split}
\min_{\mathbf{H}, F} \mathcal{Q} &= \Arrowvert r\mathbf{S} - \mathbf{H} \mathbf{H}^\mathrm{T} \Arrowvert_F^2 + n \nu \mathcal{L}(\mathbf{H}, F(\mathbf{X})) \\
&= \Arrowvert r\mathbf{S} - \mathbf{H} \mathbf{H}^\mathrm{T} \Arrowvert_F^2 + n \nu  \sum_{i=1}^n \sum_{k=1}^r l(h_{ik}, f_k(\mathbf{x}_i))
\end{split}
\label{dis_obj_ori}
\end{equation}
where $n$ is the number of training samples, $\nu$ is the penalty parameter, and $\mathcal{L}(\mathbf{H}, F(\mathbf{X}))$ denotes the quantization loss. If $\mathbf{H}$ and $\mathrm{sgn}(F(\mathbf{X}))$ are the same, the objective should be zero.

Another difficulty in solving Eq. (\ref{obj}) is the existence of pairwise similarity matrix $\mathbf{S}$, which involves at least $O(n^2)$ memory usage and at least $O(n^2)$ time consumption in matrix multiplication for $n$ training samples. Square time and space complexity makes it impossible to learn with large-scale training samples. In what follows, we propose {\em Asymmetric Low-Rank Similarity Matrix Factorization}, where we introduce the product of two low-rank matrices,  $\mathbf{P} \in \mathbb{R}^{n \times l}, \mathbf{R} \in \mathbb{R}^{n \times l}(l \ll n)$, to approximate the similarity matrix $\mathbf{S} \in \mathbb{R}^{n \times n}$:
\begin{equation}
\mathbf{S} \approx \mathbf{P}\mathbf{R}^\mathrm{T}
\label{lowrank_mat}
\end{equation}
thus Eq. (\ref{dis_obj_ori}) can be rewritten as
\begin{equation}
\min_{\mathbf{H}, F} \mathcal{Q} = \Arrowvert r\mathbf{P}\mathbf{R}^\mathrm{T}- \mathbf{H} \mathbf{H}^\mathrm{T} \Arrowvert_F^2 + n \nu \sum_{k=1}^r \sum_{i=1}^n l(h_{ik}, f_k(\mathbf{x}_i))
\label{dis_obj}
\end{equation}

If $\mathbf{H}$ is fixed, we can directly regard $\sum_{i=1}^n l (h_{ik}, f_k(\mathbf{x}_i))$ as a binary classification problem. For example, kernel SVM corresponds to a kernel-based hash function. A binary classifier with high classification accuracy corresponds to a good hash function.

We propose a discrete learning procedure to optimize Eq. (\ref{dis_obj}), which is discussed below in detail, and is summarized in Figure \ref{fig:framework}. The choice of a good similarity matrix factorization and a good binary classifier is also discussed.

\subsection{Discrete Learning Procedure}

Eq. (\ref{dis_obj}) is still a nonlinear mixed-integer program involving discrete variables $\mathbf{H}$ and hash function $F$. Similar with~\cite{liu2014discrete}, we decompose Eq. (\ref{dis_obj}) into two sub-problems: $\mathbf{F}$-Subproblem
\begin{equation}
\min_{F} \mathcal{Q}_F = \sum_{k=1}^r \sum_{i=1}^n l (h_{ik}, f_k(\mathbf{x}_i))
\label{wsub}
\end{equation}
and $\mathbf{H}$-Subproblem
\begin{equation}
\min_{\mathbf{H}} \mathcal{Q}_H = \Arrowvert r\mathbf{P}\mathbf{R}^\mathrm{T} - \mathbf{H} \mathbf{H}^\mathrm{T} \Arrowvert_F^2 + n \nu \sum_{k=1}^r \sum_{i=1}^n l (h_{ik}, f_k(\mathbf{x}_i))
\label{hsub}
\end{equation}

The subproblems (\ref{wsub}) and (\ref{hsub}) are solved alternatively. In what follows, we regard Eq. (\ref{wsub}) as $r$ independent binary classification problems, and introduce a novel clustering-based algorithm to optimize (\ref{hsub}) bit by bit.

\subsubsection{$\mathbf{F}$-Subproblem}

It is clear that $\sum_{i=1}^n l (h_{ik}, f_k(\mathbf{x}_i))$ can be regarded as a binary classification problem, in which $h_{ik}$ is the label of $\mathbf{x}_i$ and $f_k(\cdot)$ is the function to learn. Each learned binary classifier involves minimizing $\sum_{i=1}^n l (h_{ik}, f_k(\mathbf{x}_i))$ for any$ k=1,2,3,...,r$. Denote $f^*_k(\cdot)$ as the learned classification function, then $F^*(\cdot)=[f^*_1(\cdot),f^*_2(\cdot),...,f^*_r(\cdot)]^\mathrm{T}$ is the optimum of $\mathbf{F}$-Subproblem.

\subsubsection{$\mathbf{H}$-Subproblem}

We propose an efficient {\em Fast Clustering based Batch Coordinate Descent (Fast C-BCD)} algorithm to optimize $\mathbf{H}$, in which $\mathbf{H}$ is learned column by column. Let $\mathbf{b}=[b_1,...,b_n]^\mathrm{T} \in \{1,-1\}^{n}$ be the $k$th column of $\mathbf{H}$, and $\mathbf{H}'$ is the matrix of $\mathbf{H}$ excluding $\mathbf{b}$. Set $\mathbf{H}'$ fixed, and let $\mathbf{Q}=r\mathbf{P}\mathbf{R}^\mathrm{T} - \mathbf{H}{'} \mathbf{H}{'} ^{\mathrm{T}}$ the Eq. (\ref{hsub}) can be rewritten as

\begin{equation}
\begin{split}
\mathcal{Q}_H(\mathbf{b}) =& \Arrowvert r\mathbf{P}\mathbf{R}^\mathrm{T} - \mathbf{H}{'} \mathbf{H}{'} ^{\mathrm{T}} - \mathbf{b}\mathbf{b}^{\mathrm{T}} \Arrowvert_F^2 \\
&+ n \nu \sum_{i=1}^n l (b_{i}, f_k(\mathbf{x}_i)) + \mathrm{const} \\
=&\Arrowvert \mathbf{Q} \Arrowvert_F^2 + \Arrowvert \mathbf{b}\mathbf{b}^\mathrm{T} \Arrowvert^2 \\
&- 2\mathbf{b}^\mathrm{T} \mathbf{Q} \mathbf{b} + n \nu \sum_{i=1}^n l (b_{i}, f_k(\mathbf{x}_i)) + \mathrm{const} \\
=& - 2\mathbf{b}^\mathrm{T} \mathbf{Q} \mathbf{b} + n \nu \sum_{i=1}^n l (b_{i}, f_k(\mathbf{x}_i)) + \mathrm{const}
\end{split}
\end{equation}

It should be noticed that $l (b_{i}, f_k(\mathbf{x}_i))$ is a unitary binary function, thus we can rewrite it as a binary linear function: $l (b_{i}, f_k(\mathbf{x}_i))=\frac{1}{2}[l (1,f_k(\mathbf{x}_i))-l (-1,f_k(\mathbf{x}_i))] b_i + \frac{1}{2}[l (1,f_k(\mathbf{x}_i))+l (-1,f_k(\mathbf{x}_i))]$. Let $\mathbf{q}=\frac{n \nu}{2}[l (1,f_k(\mathbf{x}_1))-l (-1,f_k(\mathbf{x}_1)),...,l (1,f_k(\mathbf{x}_n))-l (-1,f_k(\mathbf{x}_n))]^\mathrm{T}$, then we have $n \nu \sum_{i=1}^n l (b_{i}, f_k(\mathbf{x}_i))=\mathbf{q}^\mathrm{T} \mathbf{b} + \mathrm{const}$. 

Discarding the constant term, we arrive at the following Binary Quadratic Programming (BQP) problem:
\begin{equation}
\min_{\mathbf{b}} g(\mathbf{b}) = - 2\mathbf{b}^\mathrm{T} \mathbf{Q} \mathbf{b} + \mathbf{q}^\mathrm{T} \mathbf{b}, \mathbf{b} \in \{-1,1\}^n
\label{bcd}
\end{equation}

Optimization of (\ref{bcd}) is still intractable. Inspired by~\cite{yang2013new,kang2016column}, we transform the problem above to an efficient clustering problem.

~\cite{yang2013new} studied the following constrained BQP problem:
\begin{equation}
    \max_{\mathbf{c}} \mathbf{c}^\mathrm{T}\mathbf{Q}_0\mathbf{c} \quad \mathrm{s.t.} \mathbf{c}^\mathrm{T}\mathbf{1}=k, \mathbf{c} \in \{0,1\}^n
    \label{equ:bqp01}
\end{equation}
where the diagonal elements of $\mathbf{Q}_0$ are zero. We should transform Eq. (\ref{bcd}) to the same form as Eq. (\ref{equ:bqp01}). First of all, we add the bit-balanced constraint $\mathbf{b}^\mathrm{T} \mathbf{1}=(n\mod 2)$ to Eq. (\ref{bcd}), which is widely used in learning-based hashing:
\begin{equation}
\begin{split}
&\max_{\mathbf{b}} g'(\mathbf{b}) = 2\mathbf{b}^\mathrm{T} \mathbf{Q} \mathbf{b} - \mathbf{q}^\mathrm{T} \mathbf{b} \\
&\mathrm{s.t.} \quad \mathbf{b} \in \{-1,1\}^n, \mathbf{b}^\mathrm{T} \mathbf{1}=(n\mod 2)
\end{split}
\label{equ:bqpminus1}
\end{equation}

Second, we set the diagonal elements of $\mathbf{Q}$ to zero. Third, we transform the domain from $\{-1,1\}^n$ to $\{0,1\}^n$ by executing the transformation of $\mathbf{b}=2\mathbf{c}-1,\mathbf{c} \in \{0,1\}^n$. Finally, we rewrite the BQP problem to the form without the linear term as well as removing the constant form, and we have the following BQP problem that is equivalent to Eq. (\ref{equ:bqpminus1}):
\begin{equation}
\begin{split}
\max_{\tilde{\mathbf{c}}} \quad& \tilde{\mathbf{c}}^\mathrm{T} \mathbf{Q}_0 \tilde{\mathbf{c}} \\
\mathrm{s.t.} \quad& \tilde{\mathbf{c}} = [\mathbf{c},1]^\mathrm{T} \in \{0,1\}^{n+1}, \mathbf{c}^\mathrm{T} \mathbf{1} = \lfloor \frac{n+1}{2} \rfloor \\
&\mathbf{Q}_0 = \begin{pmatrix} 8[\mathbf{Q}-\mathrm{diag}(\mathbf{Q})] & \mathbf{q}_0 \\ \mathbf{q}_0 & 0 \end{pmatrix} \in \mathbb{R}^{n+1} \\
&\mathbf{q}_0 = -4[\mathbf{Q}-\mathrm{diag}(\mathbf{Q})]^\mathrm{T}\mathbf{1} - \mathbf{q} \\
\end{split}
\label{equ:bqpnolinear}
\end{equation}
where $\mathrm{diag}(\mathbf{Q})$ is the diagonal matrix of $\mathbf{Q}$.

As illustrated in~\cite{yang2013new}, Eq. (\ref{equ:bqpnolinear}) can be regarded as a specific clustering problem. Given a set of vectors $\mathcal{V} = \{\mathbf{v}_i, i=1,...,n,n+1\}$, we want to find a subset $\mathcal{V}_1$ of size $K=\lfloor \frac{n+1}{2} \rfloor+1$ such that $\mathbf{v}_{n+1} \in \mathcal{V}_1$ and the sum of square of the distances between the vectors in $\mathcal{V}_1$ and the clustering center is minimized. The objective can be formulated as
\begin{equation}
\min_{|\mathcal{V}_1|=K, \mathbf{v}_{n+1} \in \mathcal{V}_1} \sum_{\mathbf{u} \in \mathcal{V}_1} \Arrowvert \mathbf{u} - \frac{\sum_{\mathbf{v} \in \mathcal{V}_1} \mathbf{v}}{K} \Arrowvert^2 \quad \mathrm{s.t.} \mathcal{V}_1 \in \mathcal{V}
\label{equ:cluster}
\end{equation}

It is clear that there exists a certain $\lambda$ such that $\mathbf{Q}_0+\lambda\mathbf{I}$ is positive semidefinite, so we have a sufficently large $\lambda$ such that $\mathbf{Q}_0+\lambda\mathbf{I}=\mathbf{V}^\mathrm{T}\mathbf{V},\mathbf{V} \in \mathbb{R}^{(n+1) \times (n+1)}$. Then we rewrite the Theorem 2.4.1 in~\cite{yang2013new} to obtain the following theorem:

\begin{thm}
If there exists $\lambda$ such that $\mathbf{Q}_0+\lambda\mathbf{I}=\mathbf{V}^\mathrm{T}\mathbf{V}, \mathbf{V} \in \mathbb{R}^{n \times n}$, and $\mathbf{v}_i$ is the $i$th column of $\mathbf{V}$, then Eq. (\ref{equ:bqpnolinear}) and Eq. (\ref{equ:cluster}) are equivalent. The global optimum of Eq. (\ref{equ:bqpnolinear}), denote $\mathbf{c}^*=[c^*_1,...,c^*_n]^\mathrm{T} \in \{0,1\}^n$, and Eq. (\ref{equ:cluster}), denote $\mathcal{V}^*_1$, have the relationship such that $c^*_i=1 \Leftrightarrow \mathbf{v}_i \in \mathcal{V}^*_1, \forall i=1,...,n$. 
\label{the:cluster}
\end{thm}

~\cite{yang2013new} proposes an iterative method to approximately solve the clustering problem. However, the Cholesky decomposition is used for getting the vectors to cluster, which involves $O(n^3)$ computational complexity. So we have to discover another efficient clustering-based algorithm.

\begin{algorithm}[t]
\small
\caption{Fast Clustering-based Batch Coordinate Descent (Fast C-BCD) Algorithm}
\textbf{Input:} hashcode $\mathbf{H}$, hyper-parameter $\lambda$.

\textbf{Output:} optimal solution of $\mathbf{H}$-Subproblem (Eq. (\ref{hsub})).

\begin{algorithmic}
\FOR {$k=1,2,...,r$}
\STATE Let $\textbf{b}_0$ be the $k$th column of $\mathbf{H}$. Denote $\textbf{b}_0$ as initialization of Eq. (\ref{equ:bqpminus1}).
\STATE Set $\mathbf{c}_0 \gets \frac{1}{2}(\mathbf{b}_0+1)$ as the initialization of Eq. (\ref{equ:bqpnolinear}).
\WHILE {not converged} 
\STATE Set $\mathbf{v}_i \in \mathcal{V}_1$ for any $[\mathbf{c}_0]_i=1$;
\STATE Compute $\mathrm{sim}(\mathbf{v}_i,\mathbf{m})$ for all $i=1,2,...,n$, according to Eq. (\ref{equ:dist});
\STATE Select $\mathbf{v}_i$ as the subset $\mathcal{V}_1$, such that they are in the first $\lfloor \frac{n+1}{2} \rfloor$ maximum value of $\mathrm{sim}(\mathbf{v}_i,\mathbf{m})$;
\STATE Set $\mathbf{c}=[c_1,...,c_n]^\mathrm{T}$ such that $c_i=1$ if $\mathbf{v}_i \in \mathcal{V}_1$, and $c_i=0$ if $\mathbf{v}_i \notin \mathcal{V}_1$;
\IF {$\mathbf{c} = \mathbf{c}_0$}
\STATE \textbf{break};
\ENDIF
\STATE $\mathbf{c}_0 \gets \mathbf{c}$;
\ENDWHILE
\STATE $\mathbf{b}=2\mathbf{c}-1$;
\STATE Replace the $k$th column of $\mathbf{H}$ with $\mathbf{b}$.
\ENDFOR
\STATE Return the updated $\mathbf{H}$.
\end{algorithmic}
\label{hsub_algo}
\end{algorithm}

We denote $[\mathbf{a}]_i$ as the $i$th column of a vector $\mathbf{a}$, and $[\mathbf{A}]_{ij}$ denotes a element of $\mathbf{A}$ at the $i$th row and the $j$th column. It can be noticed that $\| \mathbf{v}_i \|^2=\lambda$ and $\mathbf{v}_i^\mathrm{T}\mathbf{v}_j=[\mathbf{Q}_0+\lambda\mathbf{I}]_{ij}$ for arbitrary $i,j=1,2,...,n+1$, thus the square of distance of $\mathbf{v}_i \in \mathcal{V}_1$ and the clustering center $\mathbf{m}=\frac{\sum_{\mathbf{v} \in \mathcal{V}_1} \mathbf{v}}{K}$ is
\begin{equation}
\begin{split}
\Arrowvert \mathbf{v}_i-\mathbf{m} \Arrowvert^2 &= \mathbf{v}_i^2 + \mathbf{m}^2 - \frac{2 \sum_{\mathbf{v} \in \mathcal{V}_1} \mathbf{v}^\mathrm{T} \mathbf{v}_i}{K} \\
&= -\frac{2}{K} \sum_{j, \mathbf{v}_j \in \mathcal{V}_1} [\mathbf{Q}_0+\lambda\mathbf{I}]_{ij} + \mathrm{const}
\end{split}
\end{equation}

Denote the similarity between $\mathbf{v}_i$ and $\mathbf{m}$ as $ \mathrm{sim}(\mathbf{v}_i, \mathbf{m})=\sum_{j, \mathbf{v}_j \in \mathcal{V}_1} [\mathbf{Q}_0+\lambda\mathbf{I}]_{ij}$. Applying the {\em Asymmetric Low-Rank Matrix Factorization} such that $\mathbf{Q}=r\mathbf{P}\mathbf{R}^\mathrm{T} - \mathbf{H}{'} \mathbf{H}{'} ^{\mathrm{T}}$, we can greatly simplify the similarity computation:
\begin{equation}
\begin{split}
\mathrm{sim}(\mathbf{v}_i, \mathbf{m}) &= 
\begin{cases}
8[\mathbf{d}_0]_i + [\mathbf{q}_0]_i - 8[\mathbf{Q}]_{ii} + \lambda & \mathbf{v}_i \in \mathcal{V}_1, i \le n \\
8[\mathbf{d}_0]_i + [\mathbf{q}_0]_i & \mathbf{v}_i \notin \mathcal{V}_1 \\
\end{cases} \\
\mathbf{d}_0&=r\mathbf{P}(\sum_{j, \mathbf{v}_j \in \mathcal{V}_1} \mathbf{R}_{j,*})^\mathrm{T}-\mathbf{H}{'} (\sum_{j, \mathbf{v}_j \in \mathcal{V}_1} \mathbf{H}{'}_{j,*})^\mathrm{T}
\end{split}
\label{equ:dist}
\end{equation}
where $\mathbf{R}_{j,*}$ and $\mathbf{H}{'}_{j,*}$ is the $j$th row of matrix $\mathbf{R}$ and $\mathbf{H}{'}$, respectively.

Thus we can compute the distance without executing the Cholesky decomposition. For a specific $\lambda$, we modify the method proposed in ~\cite{yang2013new} and get an efficient algorithm to solve Eq. (\ref{equ:bqpminus1}), which is summarized in Algorithm \ref{hsub_algo}.

\subsection{Asymmetric Low-Rank Similarity Matrix Factorization}

As discussed before, we should find two low-rank matrices $\mathbf{P} \in \mathbb{R}^{n \times l}, \mathbf{R} \in \mathbb{R}^{n \times l}(l \ll n)$ to approximate the similarity matrix $\mathbf{S} \in \mathbb{R}^{n \times n}$. In supervised hashing algorithms, $\mathbf{S}$ is mostly defined by semantic label. $s_{ij}=1$ if $\mathbf{x}_i$ and $\mathbf{x}_j$ share the same semantic labels, and $s_{ij}=-1$ otherwise. Suppose there are $l$ semantic labels. Denote $\mathbf{y}_i = [y_{i1},y_{i2},...,y_{il}]^\mathrm{T} \in \{0,1\}^l$ as the label vector of data $\mathbf{x}_i$, in which $y_{ik}=1$ if the label of $\mathbf{x}_i$ is $k$ and $y_{ik}=0$ otherwise. Define $\mathbf{Y}=[\mathbf{y}_1,\mathbf{y}_2,...,\mathbf{y}_n]^\mathrm{T}$ and 
\begin{equation}
\mathbf{P} = [2\mathbf{Y}; \mathbf{1}_{n \time 1}] \in \mathbb{R}^{n \times (l+1)}, \mathbf{R} = [\mathbf{Y}; -\mathbf{1}_{n \time 1}] \in \mathbb{R}^{n \times (l+1)}
\label{factor}
\end{equation}
then we have the following matrix factorization:
\begin{equation}
\mathbf{P} \mathbf{R}^\mathrm{T} = 2 \mathbf{Y} \mathbf{Y}^\mathrm{T} - \mathbf{1}_{n \times n} = \mathbf{S}
\label{low_rank}
\end{equation}

It should be noticed that $\mathbf{Y}$ can be stored by sparse matrix, so just $O(n)$ space can contain all the information of $\mathbf{S}$. We can also apply Eq. (\ref{low_rank}) to existing pairwise supervised hashing methods including KSH~\cite{liu2012supervised}, TSH~\cite{lin2013general}, etc. Moreover, the {\em Asymmetric Low-Rank Similarity Matrix Factorization} scheme can be applied to any similarity matrix. For example, a variety of methods have discussed the low-rank approximation of Gaussian RBF kernel matrix~\cite{liu2011hashing,zhang2008improved}.

\subsection{Choosing Hash Functions and Initialization}
\label{k_hash_func}

For most binary classifiers, they contain certain function that predicts 1 or -1 for given data. Learning a good hash function corresponds to training a good binary classifier. There are two kinds of widely used hash functions: kernel based and deep neural network based, which is illustrated in Figure \ref{fig:framework}.

Since Eq. (\ref{dis_obj}) is a mixed-integer non-convex problem, a good initial point is very important. We can choose the existing efficient and scalable hashing algorithm to get the initialization value of $\mathbf{H}$ and hash function $F$. Different hash functions have different initialization strategies. The strategy of choosing hash function and initialization is discussed in detail at Section \ref{sec:deephashing} and \ref{sec:kernelhash}.

\subsection{Analysis}

\begin{algorithm}[t]
\small
\caption{Optimization of Discrete Supervised Hashing(DISH) Framework}
\textbf{Input:} Training data $\{ \mathbf{x}_i, y_i \}_{i=1}^n$; code length $r$; a certain hash function $F$ and a binary classification loss function $\mathcal{L} (h_{ik}, \mathbf{w}_k^\mathrm{T} k(\mathbf{x}_i))$; max iteration number $t$.

\textbf{Output:} Hash function $\mathbf{h}=\mathrm{sgn}(F(\mathbf{x}))$

\begin{enumerate}
    \item Initialize hash function $F(\cdot)$ with a certain efficient and scalabel hashing algorithm. Set $\mathbf{H}=\mathrm{sgn} (F(\mathbf{X}))$.
    \item Loop until converge or reach maximum iterations:
    \begin{itemize}
        \item \textbf{H-Subproblem:} optimize Eq. (\ref{hsub}) by {\em Clustering-based Batch Coordinate Descent(C-BCD)} algorithm proposed in Algorithm \ref{hsub_algo}.
        \item \textbf{F-Subproblem:} optimize Eq. (\ref{wsub}) by solving $r$ binary classification problems.
    \end{itemize}
\end{enumerate}
\label{dksh_algo}
\end{algorithm}

The proposed {\em Discrete Supervised Hashing(DISH) Framework} is summarized in Algorithm \ref{dksh_algo}. Denote $p$ as average count of nonzeros in each row of $\mathbf{P}$ and $\mathbf{R}$. Solving $\mathbf{H}$-Subproblem involves executing Fast C-BCD algorithm $r$ times, which will cost at most $O(2(p+r+1)T_{iter}nr)$ time, where $T_{iter}$ is the maximum number of iteration in Fast C-BCD algorithm. The space complexity is just $O(2pn+rn)$ for solving $\mathbf{H}$-Subproblem. For $\mathbf{F}$-Subproblem, $r$ binary classification problems should be solved, and many efficient classification algorithm with $O(n)$ time and space complexity can be used. Therefore, our algorithm is expected to be scalable.

\section{Discrete Learning with Deep Hashing}
\label{sec:deephashing}

By applying the deep neural networks to our DISH framework, we are able to train neural nets with the discrete constraints preserved. Denote $\Phi (\mathbf{x}_i) \in \mathbb{R}^m$ as the activation of the last hidden layer for a given image $\mathbf{x}_i$, and $\mathbf{W} = [\mathbf{w}_1,...,\mathbf{w}_r] \in \mathbb{R}^{m \times r}$ as the weight of the last fully-collected layer, where $r$ is the hash codes length. Then the activation of the output hashing layer is $F(\mathbf{x}_i)=\mathbf{W}^\mathrm{T} \Phi (\mathbf{x}_i)$ and the hash function is defined as
\begin{equation}
\mathbf{h}_i=\mathrm{sgn}(\mathbf{W}^\mathrm{T} \Phi (\mathbf{x}_i))
\end{equation}

And we use the squared hinge-loss for binary classification loss function:
\begin{equation}
\mathcal{Q}_F=\sum_{i=1}^n \sum_{k=1}^r [\max (1-h_{ik} \mathbf{w}^\mathrm{T}_k \Phi (\mathbf{x}_i))]^2
\end{equation}

There are two approaches to learn better network and avoid overfitting. First of all, if the class label is provided, we use the similar approach as CNNH+~\cite{xia2014supervised}, in which the softmax loss layer can be added above the last hidden layer:
\begin{equation}
\sum_{i=1}^n[\mathrm{softmax}(y_i, \mathbf{W}^\mathrm{T}_1 \Phi(\mathbf{x}_i)) + \mu \sum_{k=1}^r [\max (1-h_{ik} \mathbf{w}^\mathrm{T}_k \Phi (\mathbf{x}_i))]^2]
\end{equation}
where $\mu$ is a hyper-parameter, $y_i$ is the class label of $\mathbf{x}_i$, $\mathbf{W}_1$ is the parameter of fully-collected layer connecting to the softmax loss layer, and $\mathrm{softmax}(\cdot, \cdot)$ is the softmax loss.

Second, recent work show that {\em DisturbLabel}~\cite{xie2016disturblabel}, which randomly replaces part of labels to incorrect value during iteration, can prevent the network training from overfitting. Inspired by this, we can randomly flip some bits of the binary codes with probability $\alpha$ during the training of the network. We name this process as {\em DisturbBinaryCodes}, which we expect to improve the quality of binary codes.

For labeled data, we adopt the same procedure of CNNBH~\cite{guo2016hash} as initialization. CNNBH binarizes the activation of a fully-connected layer at threshold 0 to generate hashcodes, which is easy to train and achieves good performance.

\section{Discrete Learning with Kernel-Based Hashing}
\label{sec:kernelhash}

Kernel methods can embed input data to more separable space, which is widely used in machine learning algorithms such SVM, Gaussian Process, etc. As in Supervised Hashing with Kernels(KSH)~\cite{liu2012supervised} and Supervised Discrete Hashing(SDH)~\cite{Shen_2015_CVPR}, we define hash function as follows:
\begin{equation}
h_{ik}=f_k(\mathbf{x}_i)=\mathrm{sgn}(\mathbf{w}_k^\mathrm{T} k(\mathbf{x}_i))
\label{equ:kernelf}
\end{equation}
where $k(\mathbf{x})$ is the feature vector in the kernel space. $k(\mathbf{x})$ is defined as $k(\mathbf{x}) = [\phi(\mathbf{x}, \mathbf{x}_{(1)}) - \frac{1}{n} \sum_{i=1}^n  \phi(\mathbf{x}_i, \mathbf{x}_{(1)}), ..., \phi(\mathbf{x}, \mathbf{x}_{(m)}) - \frac{1}{n} \sum_{i=1}^n \phi(\mathbf{x}_i, \mathbf{x}_{(m)})]^\mathrm{T}$, where $\phi(\mathbf{x}_i, \mathbf{x}_{(j)})$ is the kernel function between $\mathbf{x}_i$ and $\mathbf{x}_{(j)}$, and $\mathbf{x}_{(j)}, j = 1,2,...,m$ are anchors. The subtraction of $\frac{1}{n} \sum_{i=1}^n \phi(\mathbf{x}_i, \mathbf{x}_{(j)})$ is to centerize the feature vectors in kernel space, so that each bit of hashcode can be more balanced. Denote $\mathbf{W}=[\mathbf{w}_1,...,\mathbf{w}_r]$ and $K(\mathbf{X})=[k(\mathbf{x}_1),...,k(\mathbf{x}_n)]^\mathrm{T}$, the hashcodes can be formulated as  $\mathbf{H} = F(\mathbf{X}) = \mathrm{sgn}(K(\mathbf{X})\mathbf{W})$, and Eq. (\ref{dis_obj}) can be rewritten as
\begin{equation}
\min_{\mathbf{H}, \mathbf{W}} \mathcal{Q} = \Arrowvert r\mathbf{P}\mathbf{R}^\mathrm{T}- \mathbf{H} \mathbf{H}^\mathrm{T} \Arrowvert_F^2 + n \nu \sum_{k=1}^r \sum_{i=1}^n l (h_{ik}, \mathbf{w}_k^\mathrm{T} k(\mathbf{x}_i))
\label{ksh_dis_obj}
\end{equation}
and we can derive three types of binary classifiers: Linear Regression, SVM and Logistic Regression, each of which corresponds to a kind of loss function $\sum_{i=1}^n l (h_{ik}, \mathbf{w}_k^\mathrm{T} k(\mathbf{x}_i))$.

For initialization, we proposed a relaxed method similar to ~\cite{liu2012supervised}, in which we remove the $\mathrm{sgn}$ function of Eq. (\ref{obj}), and optimize each column of $\mathbf{W} = [\mathbf{w}_1,\mathbf{w}_2,...,\mathbf{w}_r]$ successively.

Denote $\mathbf{H}_k$ as the first $k$ column of $\mathbf{H}$. The optimization of $\mathbf{w}_k$ is a constrained quadratic problem:
\begin{equation}
\setlength{\abovedisplayskip}{3pt}
\setlength{\belowdisplayskip}{3pt}
\begin{split}
\max_{\mathbf{w}_k} (K(\mathbf{X}) \mathbf{w}_k)^{\mathrm{T}} (\mathbf{P}\mathbf{R}^\mathrm{T} - \mathbf{H}_k \mathbf{H}_k ^{\mathrm{T}})(K(\mathbf{X}) \mathbf{w}_k) \\
\mathrm{s.t.} (K(\mathbf{X}) \mathbf{w}_k)^{\mathrm{T}} (K(\mathbf{X}) \mathbf{w}_k) = n
\end{split}
\label{spectral_obj}
\end{equation}
where $n$ is the number of training samples. Eq. (\ref{spectral_obj}) is a generalized eigenvalue problem
\begin{equation}
K(\mathbf{X})^\mathrm{T}(\mathbf{P}\mathbf{R}^\mathrm{T} - \mathbf{H}_k \mathbf{H}_k ^{\mathrm{T}})K(\mathbf{X}) \mathbf{w}_k = \lambda K(\mathbf{X})^\mathrm{T}K(\mathbf{X}) \mathbf{w}_k
\label{eigen_obj}
\end{equation}

After optimizing $\mathbf{w}_k$, denoted as $\mathbf{w}_k^0$, the $k$th column of $\mathbf{H}$ can be generated as $\mathrm{sgn}(K(\mathbf{X}) \mathbf{w}_k^0)$. The time and space complexity of the initialization procedure is just $O(n)$ because the asymmetric low-rank matrix factorization is involved.

\section{Experiments}
\label{sec:exp}

In this section, we run large-scale image retrieval experiments on three benchmarks: CIFAR-10\footnote{http://www.cs.toronto.edu/\textasciitilde kriz/cifar.html}, ImageNet\footnote{http://www.image-net.org/}~\cite{deng2009imagenet} and Nuswide\footnote{http://lms.comp.nus.edu.sg/research/NUS-WIDE.htm}~\cite{chua2009nus}. CIFAR-10 consists of 60,000 $32 \times 32$ color images from 10 object categories. ImageNet dataset is obtained from ILSVRC2012 dataset, which contains more than 1.2 million training images of 1,000 categories in total, together with 50,000 validation images. Nuswide dataset contains about 270K images collected from Flickr, and about 220K images are available from the Internet now. It associates with 81 ground truth concept labels, and each image contains multiple semantic labels. Following~\cite{liu2011hashing}, we only use the images associated with the 21 most frequent concept tags, where the total number of images is about 190K, and number of images associated with each tag is at least 5,000. 

The experimental protocols is similar with~\cite{Shen_2015_CVPR,xia2014supervised}. In CIFAR-10 dataset, we randomly select 1,000 images (100 images per class) as query set, and the rest 59,000 images as retrieval database. In Nuswide dataset, we randomly select 2,100 images (100 images per class) as the query set. And in ImageNet dataset, the provided training set are used for retrieval database, and 50,000 validation images for the query set. For CIFAR-10 and ImageNet dataset, similar images share the same semantic label. For Nuswide dataset, similar images share at least one semantic label.

For the proposed DISH framework, we name \textbf{DISH-D} and \textbf{DISH-K} as deep hashing method and kernel-based hashing method, respectively. We compare them with some recent state-of-the-art algorithms including four supervised methods: SDH~\cite{Shen_2015_CVPR}, KSH~\cite{liu2012supervised}, FastH~\cite{lin2014fast}, CCA-ITQ~\cite{gong2013iterative}, three unsupervised methods: PCA-ITQ~\cite{gong2013iterative}, AGH~\cite{liu2011hashing}, DGH~\cite{liu2014discrete}, and five deep hashing methods: CNNH+~\cite{xia2014supervised}, SFHC~\cite{lai2015simultaneous}, CNNBH~\cite{guo2016hash}, DHN~\cite{zhu2016deep}, FTDE~\cite{zhuang2016fast}. Most codes and suggested parameters of these methods are available from the corresponding authors.

Similar with~\cite{liu2012supervised,xia2014supervised}, for each dataset, we report the compared results in terms of {\em mean average precision}(MAP), precision of Hamming distance within 2, precision of top returned candidates. For Nuswide, we calculate the MAP value within the top 5000 returned neighbors, and we report the MAP of all retrieved samples on CIFAR-10 and ImageNet dataset. Groundtruths are defined by whether two candidates are similar. Scalability and sensitivity of parameters of the proposed framework will also be discussed in the corresponding subsection. The training is done on a server with Intel(R) Xeon(R) E5-2678 v3@2.50GHz CPU, 64GB RAM and a Geforce GTX TITAN X with 12GB memory.

\subsection{Experiments with Kernel-based Hashing}
\label{subsec:kernel}

\begin{table}[t]
    \centering
    \footnotesize 
    \begin{tabular}{c|c|cccc|cccc|c|c}
        \hline
         & \#  & \multicolumn{4}{c|}{MAP}  & Time/s \\ 
        Method & Training & 16 bits & 32 bits & 64 bits & 96 bits & 64 bits\\ 
        \hline
        PCA-ITQ & 59000 & 0.163 & 0.169 & 0.175 & 0.178 & 9.6 \\ 
        AGH & 59000 & 0.150 & 0.148 & 0.141 & 0.137 & 4.1 \\ 
        DGH & 59000 & 0.168 & 0.173 & 0.178 & 0.180 & 28.3 \\
        \hline
        KSH & 5000 & 0.356 & 0.390 & 0.409 & 0.415 & 756 \\ 
        SDH & 5000 & 0.341 & 0.374 & 0.397 & 0.404 & 8.6 \\ 
        \textbf{DISH-K} & 5000 & \textbf{0.380} & \textbf{0.407} & \textbf{0.420} & \textbf{0.424} & 22.2 \\
        \hline
        CCA-ITQ & 59000 & 0.301 & 0.323 & 0.328 & 0.334 & 25.4 \\ 
        KSH & 59000 & 0.405 & 0.439 & 0.468 & 0.474 & 16415 \\ 
        FastH & 59000 & 0.394 & 0.427 & 0.451 & 0.457 & 1045 \\ 
        SDH & 59000 & 0.414 & 0.442 & 0.470 & 0.474 & 76.0 \\ 
        \textbf{DISH-K} & 59000 & \textbf{0.446} & \textbf{0.474} & \textbf{0.486} & \textbf{0.491} & 83.1\\ 
        \hline
    \end{tabular}
    \caption{Comparative results of kernel-based hashing methods in MAP and training time(seconds) on CIFAR-10.The results are the average of 5 trials. We use kernel-based hash function in FastH for fair comparison.}
    \label{tab:cifar10}
\end{table}

\begin{table}[t]
    \centering
    \footnotesize
    \begin{tabular}{c|cccc|c}
        \hline
         & \multicolumn{4}{c|}{MAP} & Time/s \\
        Method & 16 bits & 32 bits & 64 bits & 96 bits & 64 bits \\
        \hline
        DGH & 0.413 & 0.421 & 0.428 & 0.429 & 97.5 \\
        CCA-ITQ & 0.508 & 0.515 & 0.522 & 0.524 & 146.1 \\
        KSH & 0.508 & 0.523 & 0.530 & \textbf{0.536} & 39543 \\
        SDH & \textbf{0.516} & \textbf{0.526} & 0.531 & 0.531 & 489.1 \\
        \textbf{DISH-K} & 0.512 & 0.524 & \textbf{0.532} & 0.533 & 216.7 \\
        \hline
    \end{tabular}
    \caption{Results of various kernel-based hashing methods on Nuswide dataset. 500 dimension bag-of-words features are extracted for evaluation.}
    \label{tab:nuswide}
\end{table}

\begin{figure}[t]
    \centering
    \includegraphics[scale=0.25]{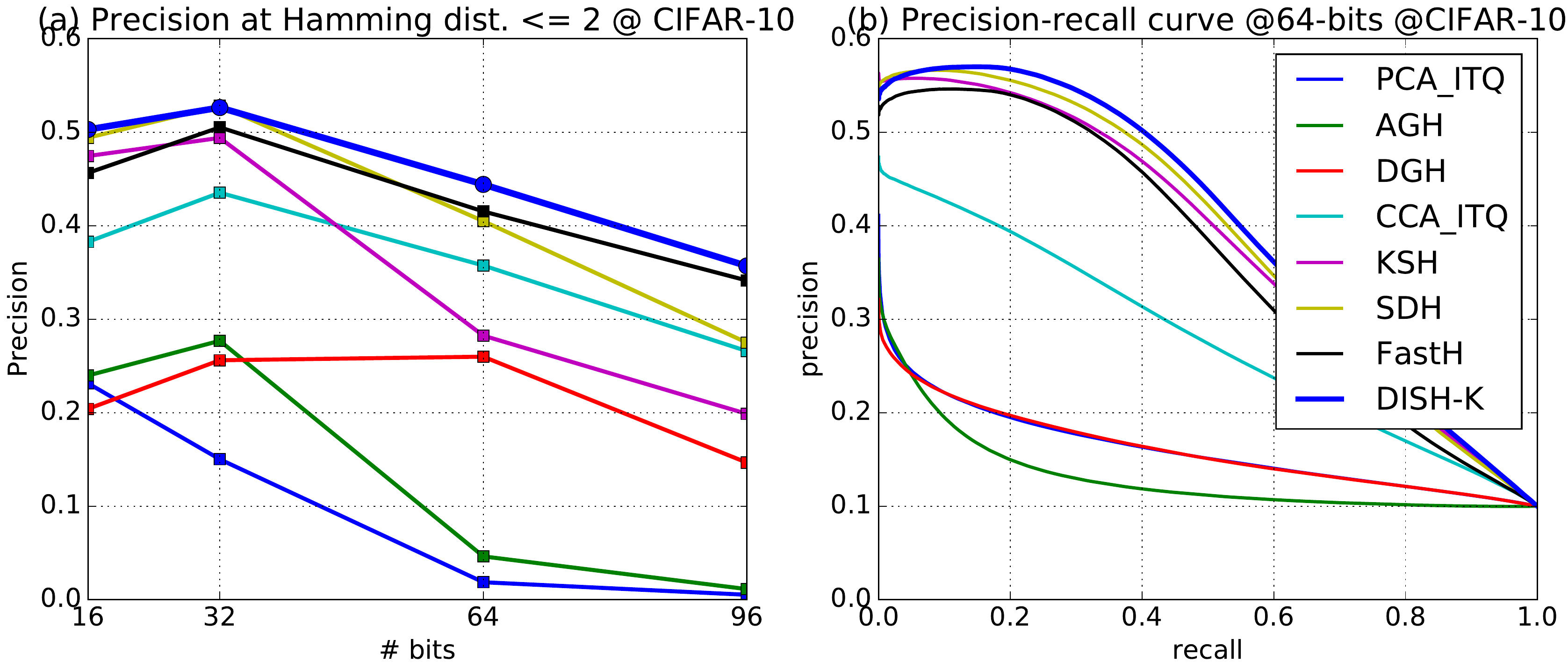}
    \caption{Precision at hamming distance within 2 value and precision-recall curve of different kernel-based methods on CIFAR-10 dataset.}
    \label{fig:cifar10}
\end{figure}

\begin{figure}[t]
    \centering
    \includegraphics[scale=0.25]{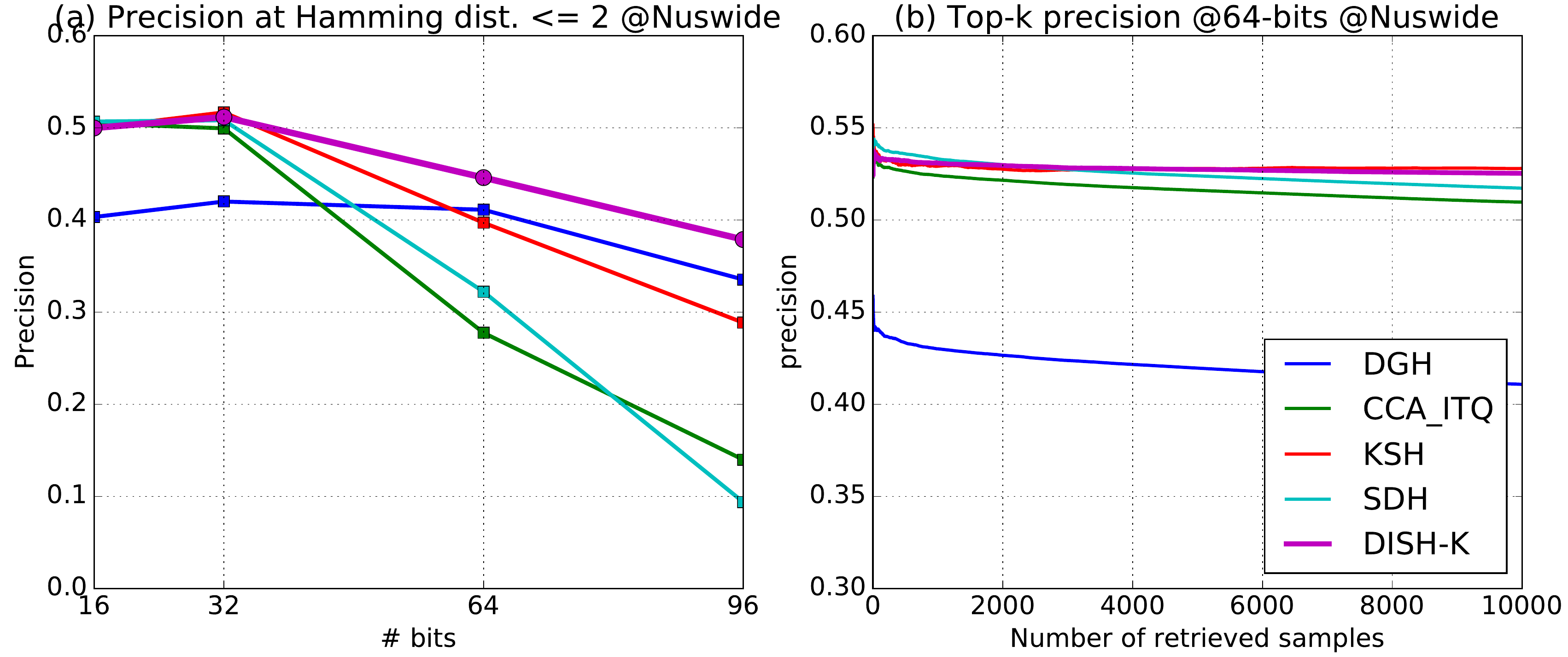}
    \caption{Precision at hamming distance within 2 value and precision-recall curve of different kernel-based methods on Nuswide dataset.}
    \label{fig:nuswide}
\end{figure}

Most existing hashing methods use the hand-crafted image features for evaluation, thus we first of all use kernel-based hash function as well as widely used hand-crafted image features to test the effectiveness of our DISH framework. For CIFAR-10 dataset, we extract 512 dimensional GIST descriptors. And for Nuswide dataset, the provided 500-dimensional bag-of-word features are used, and all features are normalized by $l_2$ norm. We use Gaussian RBF kernel $\phi(\mathbf{x}, \mathbf{y}) = \exp (-\| \mathbf{x} - \mathbf{y} \|/2\sigma^2)$ and $m=1000$ anchors. Anchors are randomly drawn from the training set and $\sigma$ is tuned for an approximate value. For FastH, we also use the kernel-based function for fair comparison. For KSH, Eq. (\ref{low_rank}) is applied to achieve faster training. For the proposed method, we set the number of iterations $t=5$ and $\nu=10^{-4}$. We solve the linear regression problem in $\mathbf{F}$-subproblem, thus the parameter $\mathbf{W}$ can be updated as follows:
\begin{equation}
\mathbf{W}^* = (K(\mathbf{X})^\mathrm{T} K(\mathbf{X}) + \lambda \mathbf{I})^{-1} K(\mathbf{X})^\mathrm{T} \mathbf{H}
\end{equation}
where we set $\lambda=0$.

Retrieval results of different methods are shown in Table \ref{tab:cifar10}, \ref{tab:nuswide} and Figure \ref{fig:cifar10}, \ref{fig:nuswide}. Our DISH method achieves best performance on CIFAR-10 dataset at both MAP and precision of Hamming distance within 2 value.\footnote{The  reported experiments use $\mathbf{H}=\mathrm{sgn}(F(\mathbf{X}))$ to generate hashcodes of the database set. Some recent works~\cite{kang2016column,zhang2014supervised} directly learn codes by similarity information for database set, and use out-of-sample technique for query set. Our DISH framework can directly adopt this protocol, in which the learned $\mathbf{H}$ can be used as database and the codes for query set can be generated by hash function $F$. The MAP value of DISH framework for CIFAR-10 is \textbf{0.724} at 64 bits, compared with 0.637 in COSDISH~\cite{kang2016column} and 0.618 in LFH~\cite{zhang2014supervised}.} And it is not surprisingly that the proposed method achieves higher precision value in longer codes, showing that discrete learning can achieve higher hash lookup success rate. Although our DISH framework is just comparable with SDH and KSH on Nuswide dataset at MAP value, the training speed of DISH is faster than SDH and KSH, showing that we can generate effective codes efficiently.

The last column in Table \ref{tab:cifar10} and \ref{tab:nuswide} show the training time. We can see that the training time of the proposed method is much faster than that of other methods involving pairwise similarity. Less than 2 minutes are consumed to train 59,000 images in our method, and it only cost 4 minutes for training nearly 200,000 data, showing that the proposed method can be easily applied to large-scale dataset.

\subsection{Experiments with Deep Hashing}
\label{subsec:deep}

\begin{table}[t]
    \centering
    \small
    \begin{tabular}{c|c|c|c|c|c|c}
    	\hline
    	layer No. & 1,2 & & 3,4 & & 5,6 & \\
        \hline
        Type & Conv & MP & Conv & MP & Conv & AP \\
        \hline
        Size & 3*3-96 & & 3*3-192 & & 3*3-192 & \\
        \hline
    \end{tabular}
    \caption{The structure of 6-layer convolutional(6conv) net. Conv means convolution layer, MP means max-pooling layer, AP means average-pooling layer. The size of pooling is 3*3 with stride 2. ReLU activation used above each convolution layer.}
    \label{tab:6conv}
\end{table}

\begin{table}[t]
    \centering
    \footnotesize
    \begin{tabular}{c|c|cccc}
        \hline
         & & \multicolumn{4}{c}{MAP} \\
        Method & Net & 12 bits & 24 bits & 32 bits & 48 bits \\
        \hline
        \multicolumn{6}{c}{without pre-training} \\
        \hline
        CNNH+ & 6conv & 0.633 & 0.625 & 0.649 & 0.654 \\
        SFHC~\cite{lai2015simultaneous} & NIN & 0.552 & 0.566 & 0.558 & 0.581 \\
        CNNBH~\cite{guo2016hash} & 6conv & 0.633 & 0.647 & 0.648 & 0.647 \\
        \textbf{DISH-D(Ours)} & 6conv & \textbf{0.667} & \textbf{0.686} & \textbf{0.690} & \textbf{0.695} \\
        \hline
        \multicolumn{6}{c}{Fine-tuning from AlexNet} \\
        \hline
        DHN~\cite{zhu2016deep} & AlexNet & 0.555 & 0.594 & 0.603 & 0.621 \\
        \textbf{DISH-D(Ours)} & AlexNet & \textbf{0.758} & \textbf{0.784} & \textbf{0.799} & \textbf{0.791} \\
        \hline
        \multicolumn{6}{c}{Fine-tuning from VGG-16 Net} \\
        \hline
        SFHC~\cite{zhuang2016fast} & VGG-16 & N/A & 0.677 & 0.688 & 0.699 \\
        FTDE~\cite{zhuang2016fast} & VGG-16 & N/A & 0.760 & 0.768 & 0.769 \\
        \textbf{DISH-D(Ours)} & VGG-16 & \textbf{0.841} & \textbf{0.854} & \textbf{0.859} & \textbf{0.857} \\
        \hline
    \end{tabular}
    \caption{Results of deep hashing methods in MAP on CIFAR-10. For this dataset, 5,000 data are randomly sampled as training set. NIN is a 8-layer net with the network-in-network structure as basic framework.}
    \label{tab:cifar10_deep}
\end{table}

\begin{table}[t]
    \centering
    \footnotesize
    \begin{tabular}{c|c|cccc}
        \hline
         & & \multicolumn{4}{c}{MAP} \\
        Method & Net & 12 bits & 24 bits & 32 bits & 48 bits \\
        \hline
        \multicolumn{6}{c}{Fine-tuning from AlexNet} \\
        \hline
        DHN~\cite{zhu2016deep} & AlexNet & 0.708 & 0.735 & 0.748 & 0.758 \\
        \textbf{DISH-D(Ours)} & AlexNet & \textbf{0.787} & \textbf{0.810} & \textbf{0.810} & \textbf{0.813} \\
        \hline
        \multicolumn{6}{c}{Fine-tuning from VGG-16 Net} \\
        \hline
        SFHC~\cite{zhuang2016fast} & VGG-16 & N/A & 0.718 & 0.720 & 0.723 \\
        FTDE~\cite{zhuang2016fast} & VGG-16 & N/A & 0.750 & 0.756 & 0.760 \\
        \textbf{DISH-D(Ours)} & VGG-16 & \textbf{0.833} & \textbf{0.850} & \textbf{0.850} & \textbf{0.856} \\
        \hline
    \end{tabular}
    \caption{Results of deep hashing methods in MAP on Nuswide dataset.}
    \label{tab:nuswide_deep}
\end{table}

\begin{figure}[t]
    \centering
    \includegraphics[scale=0.25]{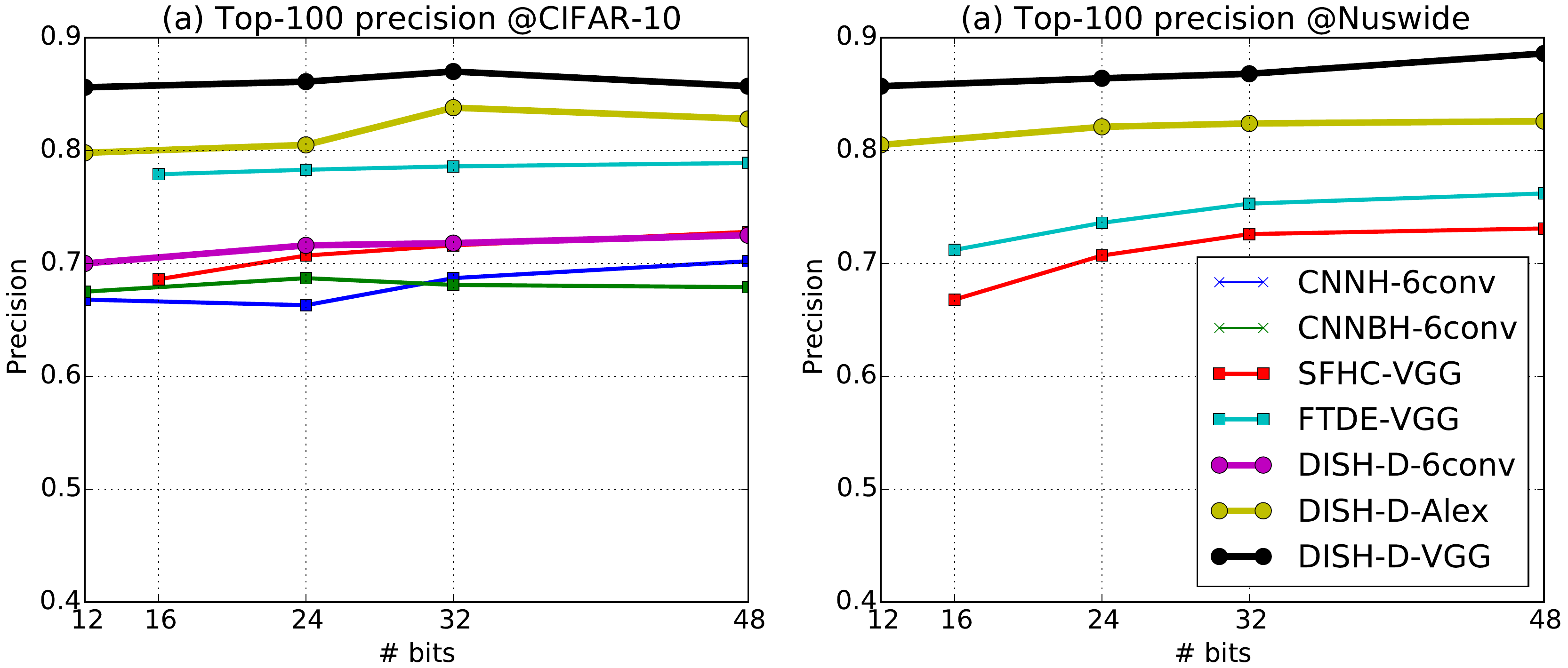}
    \caption{Top-100 precision of different deep hashing methods on CIFAR-10 and Nuswide dataset.}
    \label{fig:deephash}
\end{figure}

\subsubsection{Network Structure and Experimental Setup}

Below the output hashing layer, the network contains multiple convolutional, pooling and fully-connected layers. Different models can lead to significantly different retrieval performance. For fair comparison, we use three types of model for evaluation: a 6-layer convolutional network(6conv) which is shown in Table \ref{tab:6conv}; the pre-trained AlexNet~\cite{krizhevsky2012imagenet}; and the pre-trained VGG-16 net~\cite{simonyan2014very}.

We evaluate the deep hashing methods on CIFAR-10 and Nuswide dataset. For CIFAR-10, following~\cite{xia2014supervised}, we random sample 5,000 images (500 images per class) as training set. For Nuswide, we use the database images as training images. We resize images to $256 \times 256$ to train AlexNet and VGG-16 net.

We implement the proposed model based on the \textbf{Caffe}~\cite{jia2014caffe} framework. We set the number of iterations $t=3$ and $\nu=10^{-4}$. For 6conv model, the initial learning rate is set to $0.01$. For AlexNet and VGG-16 model, the weight before the last hidden layer is copied from pre-trained model, the initial learning rate is $0.001$ and the learning rate of last fully-connected layer is 10 times that of lower layers. The base learning rate drops 50\% after each iteration. We use stochastic gradient descent (SGD) with momentum $0.9$ and the weight decay is set to $0.0005$. The parameter of {\em DisturbBinaryCodes} is set to $\alpha=0.3$ for CIFAR-10 dataset to avoid over-fitting. It is selected with cross-validation.

\subsubsection{Comparison with State-of-the-art}
Table \ref{tab:cifar10_deep}, \ref{tab:nuswide_deep} and Figure \ref{fig:deephash} shows the retrieval performance on existing deep hashing methods. Note that the results with citation are copied from the corresponding papers. On a variety of network structures, our DISH framework achieves much better MAP and precision of Hamming distance within 2 value by a margin of \textbf{4\%-10\%}. Compared with DHN and CNNBH, it can be seen clearly that combining discrete learning with deep hashing can greatly improve the performance compared with relaxed-based deep hashing methods. Although the network structure of the proposed framework is similar with CNNH+ and FTDE, our DISH framework is much better than the latter. It is likely that the codes generated by discrete learning procedure in DISH framework can embed more information of the distribution of data.

\begin{table}[t]
    \centering
    \footnotesize
    \begin{tabular}{c|c|cc}
        \hline
         & & \multicolumn{2}{c}{Training time(hours)} \\
        Method & Net & CIFAR-10 & Nuswide \\
        \hline
        SFHC~\cite{zhuang2016fast} & VGG-16 & 174 & 365 \\
        FTDE~\cite{zhuang2016fast} & VGG-16 & 15 & 32 \\
        \textbf{DISH-D(Ours)} & VGG-16 & \textbf{4} & \textbf{9} \\
        \hline
    \end{tabular}
    \caption{Training time of various deep hashing methods. VGG-16 net is used for evaluation. Our DISH framework performs much faster than others.}
    \label{tab:deep_time}
\end{table}

Table \ref{tab:deep_time} summarize the training time of some state-of-the-art methods. VGG-16 net is used for evaluation. As expected, the training speed is much faster than triplet-based deep hashing methods. It takes less than 1 day to generate efficient binary codes by VGG-16 net, thus we can also train binary codes efficiently with deep neural nets.

\subsection{Scalability: Training Hashcodes on ImageNet}

\begin{table}[t]
    \centering
    \footnotesize
    \begin{tabular}{c|cccc|c}
        \hline
         & \multicolumn{4}{c|}{MAP} & Time/s \\
        Method & 32 bits & 64 bits & 128 bits & 256 bits & 128 bits \\
        \hline
        DGH & 0.044 & 0.077 & 0.108 & 0.135 & 121.5 \\
        CCA-ITQ & 0.096 & 0.183 & 0.271 & 0.340 & 421.0 \\
        KSH & 0.131 & 0.219 & 0.282 & 0.319 & 9686 \\
        SDH & 0.133 & 0.219 & 0.284 & 0.325 & 5831 \\
        \textbf{DISH-K} & \textbf{0.176} & \textbf{0.250} & \textbf{0.313} & \textbf{0.357} & 257.3 \\
        \hline
        $l_2$ distance & \multicolumn{4}{c|}{0.306} & \\
        \hline
        FastH-Full~\cite{lin2014fast} & \multicolumn{4}{c|}{0.390(128 bits)} & N/A \\
        \textbf{DISH-K-Full} & \multicolumn{4}{c|}{\textbf{0.417}(128 bits)} & 2050 \\
        \hline
    \end{tabular}
    \caption{Comparative results of various methods on ImageNet dataset. For the first five rows, 100,000 samples are used for training. FastH-Full and DISH-K-Full used all 1.2 million training samples at 128 bits.}
    \label{tab:imagenet}
\end{table}

We evaluate with the ImageNet dataset for testing the scalability of our proposed framework. We use the provided training set as the retrieval database and 50,000 validation set as the query set. For kernel-based methods, 4096-dimensional features are extracted from the last hidden layer of VGG-19 net~\cite{simonyan2014very}. We subtract the mean of training image features, followed by normalizing the feature representation with $l_2$-norm. We found that it may not converge during training because the number of dissimilar pairs is much larger than the similar ones. To tackle the problem, we set the value of similar pairs as 9 in $\mathbf{S}$ to balance the dissimilar pairs. Thus the similarity matrix can be represented as $\mathbf{S}=10\mathbf{Y}\mathbf{Y}^\mathrm{T}-\mathbf{1}$, where $\mathbf{Y}$ is defined the same as Eq. (\ref{low_rank}).

\begin{figure}[t]
    \centering
    \includegraphics[scale=0.253]{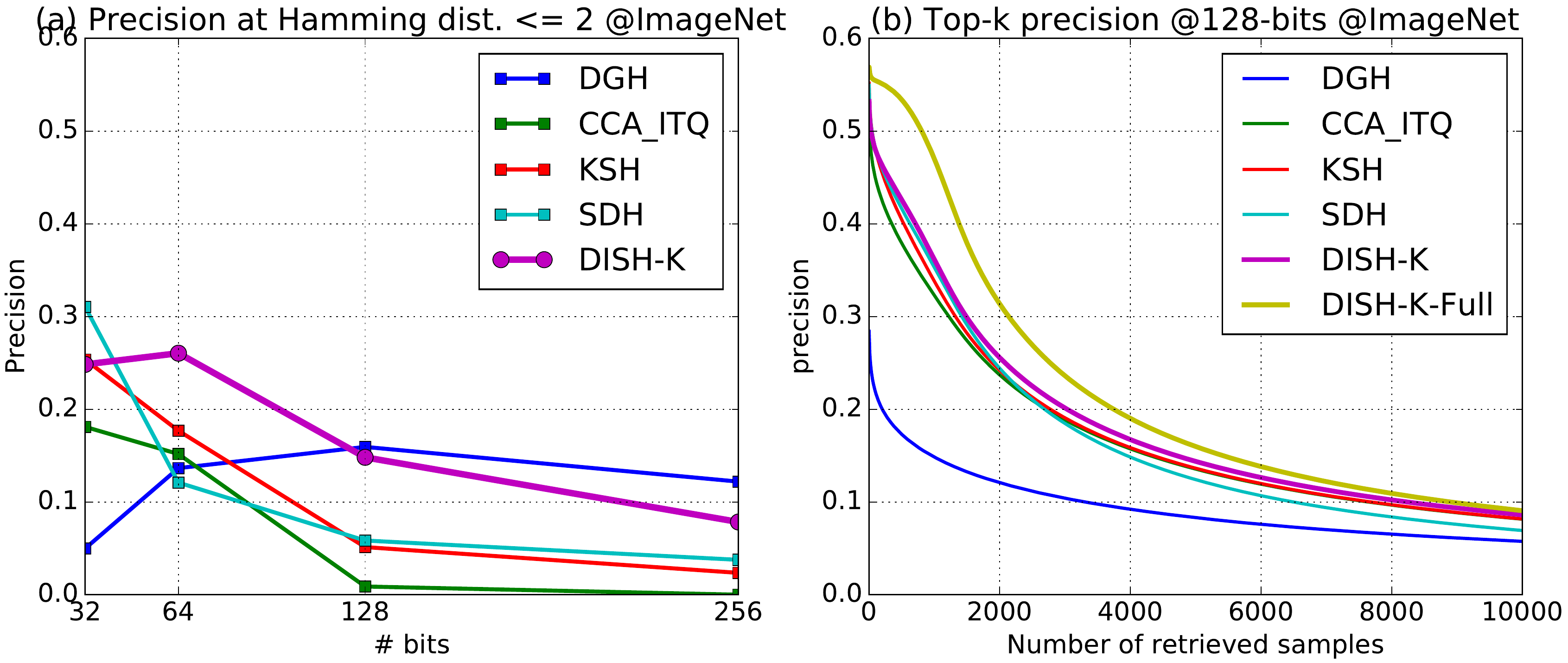}
    \caption{Results of precision at Hamming distance within 2 and Top-k precision of different methods on ImageNet dataset.}
    \label{fig:imagenet}
\end{figure}

First, we sample 100,000 images (100 samples per category) as the training set. The results are shown in Table \ref{tab:imagenet} and Figure \ref{fig:imagenet}. Retrieval results based on $l_2$ distance of 4096-dimensional features are also reported. Similar with results on CIFAR-10, the proposed DISH achieves the best performance, especially on MAP($\sim0.03$). DGH algorithm gets higher precision of Hamming distance within 2 in longer codes, but has much poorer MAP. Overall, the performances of DISH and DGH show the power of discrete learning methods.

To test the scalability of proposed framework, we train hash function with the full 1.2 million ImageNet dataset. It takes less than 1 hour for training, showing the framework can easily applied to large-scale dataset. DISH-K-Full in Table \ref{tab:imagenet} and Figure \ref{fig:imagenet} shows the result. The MAP value improves over \textbf{33\%} if the whole training data are used, and it is shown clearly that DISH method is able to deal with millions of data.

It is interesting that the proposed method outperforms the method based on $l_2$ distance, showing that DISH can not only reduce the dimensionality of data, but also embed more useful information.

\begin{figure}[t]
    \centering
    \includegraphics[scale=0.254]{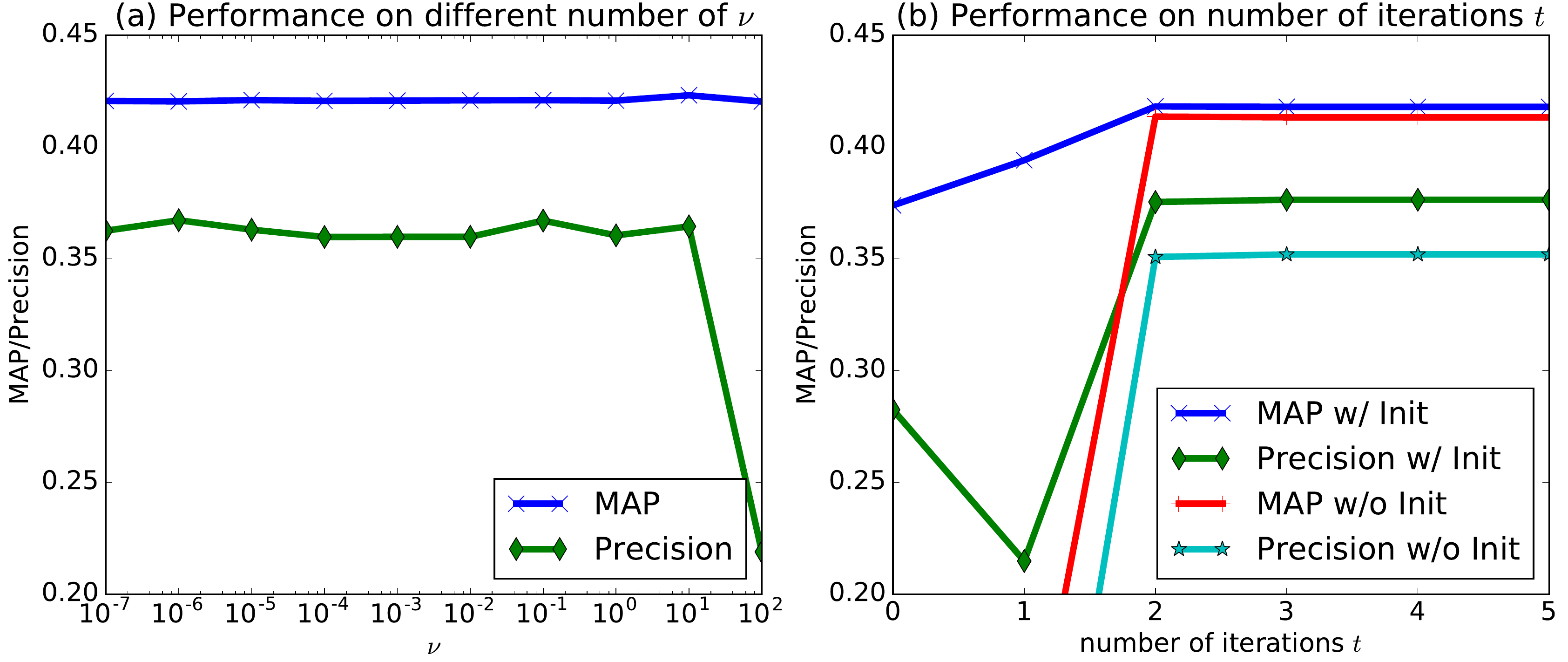}
    \caption{Comparative results of different $\nu$ and different number of iterations $t$. "w/ Init" means we perform initialization before discrete optimization and "w/o Init" otherwise. \textbf{DISH-K} is used for evaluation.}
    \label{fig:nu}
    \label{fig:iteration}
\end{figure}

\begin{table}[t]
    \centering
    \footnotesize
    \begin{tabular}{c|cccccc}
        \hline
        $\alpha$ & 0 & 0.1 & 0.2 & 0.3 & 0.4 & 0.5 \\
        \hline
        MAP & 0.687 & 0.681 & 0.699 & 0.699 & 0.706 & 0.326 \\
        Precision & 0.688 & 0.676 & 0.678 & 0.673 & 0.647 & 0.001 \\
        \hline
    \end{tabular}
    \caption{Results of various {\em DisturbBinaryCodes} ratio $\alpha$. 6conv net is used for evaluation.}
    \label{tab:disturb}
\end{table}

\subsection{Sensitivity to Parameters}
\label{subsec:discussion}

In this subsection, the efficiency on different settings of the DISH framework (DISH-K and DISH-D) is evaluated. We test our method on CIFAR-10 dataset, and the experimental settings are the same as Section \ref{subsec:deep} and \ref{subsec:kernel}. 5,000 samples (500 samples per class) from the retrieval database are used for training set. The code length is 64.

\subsubsection{Influence of $\nu$}
Figure \ref{fig:nu}(a) shows the performance on different values of $\nu$. It is shown that the algorithm is not sensitive to $\nu$ over a wide range, so we can choose this parameter freely.

\subsubsection{Influence of initialization and discrete learning procedure}
Figure \ref{fig:iteration}(b) shows the performance on the existence of initialization and different number of iterations $t$. "w/ Init" means we perform initialization before discrete optimization, and "w/o Init" otherwise. $t=0$ means we do not perform discrete optimization. It is clear that both initialization and discrete learning procedure makes the retrieval performance much better, especially for precision of Hamming distance within 2. The performance will not be improved after just a few iterations, showing our method can converge fast.

\subsubsection{Influence of {\em DisturbBinaryCodes}}

Table \ref{tab:disturb} shows the performance on different disturb rate $\alpha$, $\alpha=0$ means no {\em DistrubBinaryCodes} is involved. It is seen clearly that the performance can be greatly improved if a proper $\alpha$ is set, showing that {\em DistrubBinaryCodes} can be performed as a regularizer on the loss-layer and avoid over-fitting.

\section{Conclusion}
\label{sec:conclusion}

In this paper, we propose a novel discrete supervised hashing framework for supervised hashing problem. To deal with the discrete constraints in the binary codes, a discrete learning procedure is proposed to learn binary codes directly. We decompose the learning procedure into two sub-problems: one is to learn binary codes directly, the other is solving several binary classification problems to learn hash functions. To leverage the pairwise similarity and reduce the training time simultaneously, a novel {\em Asymmetric Low-rank Similarity Matrix Factorization} approach is introduced, and we propose the {\em Fast Clustering-based Batch Coordinate Descent} method to learn binary codes efficiently. The DISH framework can be easily adopted to arbitrary binary classifier, including deep neural networks and kernel-based methods. Experimental results on large-scale datasets demonstrate the efficiency and scalability of our DISH framework.

\section*{Acknowledgment}

This work was supported by the National Basic Research Program (973 Program) of China (No. 2013CB329403), and the National Natural Science Foundation of China (Nos. 61332007, 91420201 and 61620106010).

\bibliographystyle{unsrt}
\bibliography{references}

\begin{thebibliography}{10}

\bibitem{li2011hashing}
Ping Li, Anshumali Shrivastava, Joshua~L Moore, and Arnd~C K{\"o}nig.
\newblock Hashing algorithms for large-scale learning.
\newblock In {\em Advances in neural information processing systems}, pages
  2672--2680, 2011.

\bibitem{liu2012supervised}
Wei Liu, Jun Wang, Rongrong Ji, Yu-Gang Jiang, and Shih-Fu Chang.
\newblock Supervised hashing with kernels.
\newblock In {\em Computer Vision and Pattern Recognition (CVPR), 2012 IEEE
  Conference on}, pages 2074--2081. IEEE, 2012.

\bibitem{li2013sign}
Ping Li, Gennady Samorodnitsk, and John Hopcroft.
\newblock Sign cauchy projections and chi-square kernel.
\newblock In {\em Advances in Neural Information Processing Systems}, pages
  2571--2579, 2013.

\bibitem{datar2004locality}
Mayur Datar, Nicole Immorlica, Piotr Indyk, and Vahab~S Mirrokni.
\newblock Locality-sensitive hashing scheme based on p-stable distributions.
\newblock In {\em Proceedings of the twentieth annual symposium on
  Computational geometry}, pages 253--262. ACM, 2004.

\bibitem{gionis1999similarity}
Aristides Gionis, Piotr Indyk, Rajeev Motwani, et~al.
\newblock Similarity search in high dimensions via hashing.
\newblock In {\em VLDB}, volume~99, pages 518--529, 1999.

\bibitem{broder1998min}
Andrei~Z Broder, Moses Charikar, Alan~M Frieze, and Michael Mitzenmacher.
\newblock Min-wise independent permutations.
\newblock In {\em Proceedings of the thirtieth annual ACM symposium on Theory
  of computing}, pages 327--336. ACM, 1998.

\bibitem{weiss2009spectral}
Yair Weiss, Antonio Torralba, and Rob Fergus.
\newblock Spectral hashing.
\newblock In {\em Advances in neural information processing systems}, pages
  1753--1760, 2009.

\bibitem{gong2013iterative}
Yunchao Gong, Svetlana Lazebnik, Albert Gordo, and Florent Perronnin.
\newblock Iterative quantization: A procrustean approach to learning binary
  codes for large-scale image retrieval.
\newblock {\em Pattern Analysis and Machine Intelligence, IEEE Transactions
  on}, 35(12):2916--2929, 2013.

\bibitem{liu2011hashing}
Wei Liu, Jun Wang, Sanjiv Kumar, and Shih-Fu Chang.
\newblock Hashing with graphs.
\newblock In {\em Proceedings of the 28th international conference on machine
  learning (ICML-11)}, pages 1--8, 2011.

\bibitem{kong2012isotropic}
Weihao Kong and Wu-Jun Li.
\newblock Isotropic hashing.
\newblock In {\em Advances in Neural Information Processing Systems}, pages
  1646--1654, 2012.

\bibitem{liu2014discrete}
Wei Liu, Cun Mu, Sanjiv Kumar, and Shih-Fu Chang.
\newblock Discrete graph hashing.
\newblock In {\em Advances in Neural Information Processing Systems}, pages
  3419--3427, 2014.

\bibitem{kulis2009learning}
Brian Kulis and Trevor Darrell.
\newblock Learning to hash with binary reconstructive embeddings.
\newblock In {\em Advances in neural information processing systems}, pages
  1042--1050, 2009.

\bibitem{norouzi2011minimal}
Mohammad Norouzi and David~M Blei.
\newblock Minimal loss hashing for compact binary codes.
\newblock In {\em Proceedings of the 28th international conference on machine
  learning (ICML-11)}, pages 353--360, 2011.

\bibitem{lin2014fast}
Guosheng Lin, Chunhua Shen, Qinfeng Shi, Anton van~den Hengel, and David Suter.
\newblock Fast supervised hashing with decision trees for high-dimensional
  data.
\newblock In {\em Computer Vision and Pattern Recognition (CVPR), 2014 IEEE
  Conference on}, pages 1971--1978. IEEE, 2014.

\bibitem{Shen_2015_CVPR}
Fumin Shen, Chunhua Shen, Wei Liu, and Heng Tao~Shen.
\newblock Supervised discrete hashing.
\newblock In {\em The IEEE Conference on Computer Vision and Pattern
  Recognition (CVPR)}, June 2015.

\bibitem{deng2009imagenet}
Jia Deng, Wei Dong, Richard Socher, Li-Jia Li, Kai Li, and Li~Fei-Fei.
\newblock Imagenet: A large-scale hierarchical image database.
\newblock In {\em Computer Vision and Pattern Recognition, 2009. CVPR 2009.
  IEEE Conference on}, pages 248--255. IEEE, 2009.

\bibitem{xia2014supervised}
Rongkai Xia, Yan Pan, Hanjiang Lai, Cong Liu, and Shuicheng Yan.
\newblock Supervised hashing for image retrieval via image representation
  learning.
\newblock In {\em Proceedings of the AAAI Conference on Artificial
  Intellignece}, pages 2156--2162, 2014.

\bibitem{lai2015simultaneous}
Hanjiang Lai, Yan Pan, Ye~Liu, and Shuicheng Yan.
\newblock Simultaneous feature learning and hash coding with deep neural
  networks.
\newblock In {\em The IEEE Conference on Computer Vision and Pattern
  Recognition (CVPR)}, June 2015.

\bibitem{guo2016hash}
Jinma Guo, Shifeng Zhang, and Jianmin Li.
\newblock Hash learning with convolutional neural networks for semantic based
  image retrieval.
\newblock In {\em Advances in Knowledge Discovery and Data Mining}, pages
  227--238. Springer, 2016.

\bibitem{zhu2016deep}
Han Zhu, Mingsheng Long, Jianmin Wang, and Yue Cao.
\newblock Deep hashing network for efficient similarity retrieval.
\newblock In {\em Thirtieth AAAI Conference on Artificial Intelligence}, 2016.

\bibitem{lin2013general}
Guosheng Lin, Chunhua Shen, David Suter, and Anton van~den Hengel.
\newblock A general two-step approach to learning-based hashing.
\newblock In {\em Computer Vision (ICCV), 2013 IEEE International Conference
  on}, pages 2552--2559. IEEE, 2013.

\bibitem{krizhevsky2012imagenet}
Alex Krizhevsky, Ilya Sutskever, and Geoffrey~E Hinton.
\newblock Imagenet classification with deep convolutional neural networks.
\newblock In {\em Advances in neural information processing systems}, pages
  1097--1105, 2012.

\bibitem{simonyan2014very}
Karen Simonyan and Andrew Zisserman.
\newblock Very deep convolutional networks for large-scale image recognition.
\newblock {\em arXiv preprint arXiv:1409.1556}, 2014.

\bibitem{he2015deep}
Kaiming He, Xiangyu Zhang, Shaoqing Ren, and Jian Sun.
\newblock Deep residual learning for image recognition.
\newblock {\em arXiv preprint arXiv:1512.03385}, 2015.

\bibitem{ren2015faster}
Shaoqing Ren, Kaiming He, Ross Girshick, and Jian Sun.
\newblock Faster r-cnn: Towards real-time object detection with region proposal
  networks.
\newblock In {\em Advances in Neural Information Processing Systems}, pages
  91--99, 2015.

\bibitem{wan2014deep}
Ji~Wan, Dayong Wang, Steven Chu~Hong Hoi, Pengcheng Wu, Jianke Zhu, Yongdong
  Zhang, and Jintao Li.
\newblock Deep learning for content-based image retrieval: A comprehensive
  study.
\newblock In {\em Proceedings of the ACM International Conference on
  Multimedia}, pages 157--166. ACM, 2014.

\bibitem{schroff2015facenet}
Florian Schroff, Dmitry Kalenichenko, and James Philbin.
\newblock Facenet: A unified embedding for face recognition and clustering.
\newblock In {\em Proceedings of the IEEE Conference on Computer Vision and
  Pattern Recognition}, pages 815--823, 2015.

\bibitem{zhuang2016fast}
Bohan Zhuang, Guosheng Lin, Chunhua Shen, and Ian Reid.
\newblock Fast training of triplet-based deep binary embedding networks.
\newblock {\em arXiv preprint arXiv:1603.02844}, 2016.

\bibitem{yang2013new}
Rui Yang.
\newblock {\em New results on some quadratic programming problems}.
\newblock PhD thesis, University of Illinois at Urbana-Champaign, 2013.

\bibitem{kang2016column}
Wang-Cheng Kang, Wu-Jun Li, and Zhi-Hua Zhou.
\newblock Column sampling based discrete supervised hashing.
\newblock In {\em Proceedings of the AAAI Conference on Artificial
  Intellignece}, 2016.

\bibitem{zhang2008improved}
Kai Zhang, Ivor~W Tsang, and James~T Kwok.
\newblock Improved nystr{\"o}m low-rank approximation and error analysis.
\newblock In {\em Proceedings of the 25th international conference on Machine
  learning}, pages 1232--1239. ACM, 2008.

\bibitem{xie2016disturblabel}
Lingxi Xie, Jingdong Wang, Zhen Wei, Meng Wang, and Qi~Tian.
\newblock Disturblabel: Regularizing cnn on the loss layer.
\newblock {\em arXiv preprint arXiv:1605.00055}, 2016.

\bibitem{chua2009nus}
Tat-Seng Chua, Jinhui Tang, Richang Hong, Haojie Li, Zhiping Luo, and Yantao
  Zheng.
\newblock Nus-wide: a real-world web image database from national university of
  singapore.
\newblock In {\em Proceedings of the ACM international conference on image and
  video retrieval}, page~48. ACM, 2009.

\bibitem{zhang2014supervised}
Peichao Zhang, Wei Zhang, Wu-Jun Li, and Minyi Guo.
\newblock Supervised hashing with latent factor models.
\newblock In {\em Proceedings of the 37th international ACM SIGIR conference on
  Research \& development in information retrieval}, pages 173--182. ACM, 2014.

\bibitem{jia2014caffe}
Yangqing Jia, Evan Shelhamer, Jeff Donahue, Sergey Karayev, Jonathan Long, Ross
  Girshick, Sergio Guadarrama, and Trevor Darrell.
\newblock Caffe: Convolutional architecture for fast feature embedding.
\newblock In {\em Proceedings of the ACM International Conference on
  Multimedia}, pages 675--678. ACM, 2014.

\end{thebibliography}

\end{document}